\documentclass[journal]{IEEEtran}
\usepackage[noadjust]{cite}

\usepackage{multirow,multicol}

\usepackage[tbtags]{amsmath}
\usepackage{amsbsy}
\usepackage{amssymb}
\usepackage{amsfonts}
\usepackage{bbm}

\usepackage{graphicx}
\usepackage[caption=false]{subfig}
\usepackage{url}
\usepackage{textcomp}
\usepackage{algorithmic}
\usepackage{algorithm}
\usepackage{balance}

\usepackage{array}

\usepackage{hyperref}
\usepackage{color}

{\begin{list}               %    with flush left bullets.
    {$\bullet$ \hfill}{
        \setlength{\leftmargin}{\parindent}
        \setlength{\parsep}{0.04\baselineskip}
        \setlength{\itemsep}{0.5\parsep}
        \setlength{\labelwidth}{\leftmargin}
        \setlength{\labelsep}{0em}}
    }
{\end{list}}

\providecommand{\cref}[1]{Chapter~\ref{#1}}
\providecommand{\sref}[1]{Section~\ref{#1}}
\providecommand{\fref}[1]{Fig.~\ref{#1}}

\renewcommand{\vec}[1]{\ensuremath{\boldsymbol{#1}}}
\providecommand{\mat}[1]{\ensuremath{\boldsymbol{#1}}}

\graphicspath{{figs/}}
\DeclareGraphicsExtensions{.jpg, .png}

\begin{document}

\title{Camera-Aware Multi-Resolution Analysis for Raw Sensor Data Compression}

\author{Yeejin~Lee,~\IEEEmembership{Student Member,~IEEE,}
        Keigo~Hirakawa,~\IEEEmembership{Senior Member,~IEEE,}
        and~Truong~Q.~Nguyen,~\IEEEmembership{Fellow,~IEEE}% <-this % stops a space
\thanks{Y. Lee and T. Q. Nguyen are with the Department
of Electrical and Computer Engineering, University of California, San Diego,
CA, 92093 USA (e-mail: yel031@eng.ucsd.edu, tqn001@eng.ucsd.edu).}% <-this % stops a space
\thanks{K. Hirakawa is with the Department of Electrical and Computer Engineering, University of Dayton,
OH, 45469 USA (e-mail: khirakawa1@udayton.edu).}}% <-

% make the title area
\maketitle

% As a general rule, do not put math, special symbols or citations
% in the abstract or keywords.
\begin{abstract}
We propose a novel lossless and lossy compression scheme for color filter array~(CFA) sampled images based on the  wavelet transform of them. Our analysis suggests that the wavelet coefficients of {\it HL} and {\it LH} subbands are highly correlated. Hence, we decorrelate Mallat wavelet packet decomposition to further sparsify the coefficients. In addition, we develop a camera processing pipeline for compressing CFA sampled images aimed at maximizing the quality of the color images constructed from the compressed CFA sampled images. We validated our theoretical analysis and the performance of the proposed compression scheme using images of natural scenes captured in a raw format. The experimental results verify that our proposed method improves coding efficiency relative to the standard and the state-of-the-art compression schemes CFA sampled images.   
\end{abstract}

% Note that keywords are not normally used for peerreview papers.

\section{Introduction}
\IEEEPARstart{C}{olor} filter array~(CFA) refers to a spatial multiplexing of red, green, and blue filters over the image sensor surface. The popular Bayer CFA pattern is shown in \fref{fig:bayerpattern}. The raw sensor data captured by this configuration is therefore a subsampled version of a full-color image, where each pixel sensor measures the intensity of only a red, green, or blue value. Demosaicking process interpolates the raw image and recovers the full-color image representation. Besides demosaicking, the captured data undergo defective sensor pixel removal, color correction, gamma correction, noise suppression, etc. Once the camera processing pipeline renders the color image, an image compression algorithm transforms the processed image to a storage format that requires fewer bits.

When compressing the image data, a digital data representation of the best quality is highly desirable in many applications, such as digital cinema, broadcast, medical imaging, image archives, etc. In such applications, lossless compression is preferable so that the original data can be reconstructed perfectly. However, lossless compression often requires large storage space. Hence, lossy compression, which could significantly reduce data size for storing, processing, and transmitting, is more common in consumer applications. 

On the other hand, photographers often work directly with raw sensor data to maximize control over the post-processing. However, storing raw sensor data is difficult. With no standard method for compressing this type of data, the storage size of the raw sensor data is typically very large. At the same time, compressing the raw sensor data also has the potential to improve compression performance relative to working with a full-color image. Despite the fact that the color images requires more bits to encode all color components, compression of CFA sampled images has received very limited attention in the literature~\cite{Lin16, Lakshmi16, Trifan16, Kim14, Lee12, Malvar12, Moghadam10, Chung08, Zhang06, Lee01, Koh01}.

To this end, we propose a novel compression scheme for raw sensor data. Our work is inspired by the CFA compression method of \cite{Zhang06}, the wavelet sampling theory of \cite{Hirakawa11}, and the demosaicking work of \cite{Korneliussen14, Hirakawa07}. Specifically, the CFA sampled image compression scheme in \cite{Zhang06} leveraged a heuristic observation made about the behavior of the Mallat wavelet packet transform coefficients corresponding to the CFA sampled image. A more rigorous analysis of wavelet sampling in \cite{Hirakawa11} revealed that Mallat wavelet coefficients are highly redundant. We previously used this fact to develop demosaicking~\cite{Korneliussen14, Hirakawa07} and denoising methods~\cite{Hirakawa07}. 

The remainder of this article is organized as follows. In \sref{sec:background}, we provide theoretical analysis and review of wavelet transform for CFA sampled images. In \sref{sec:method}, we develop a scheme to further decorrelate the Mallat coefficients. We then propose a lossless and lossy compression of CFA sampled images leveraging this decorrelated Mallat wavelet structure. In addition, we design a camera processing pipeline aimed at maximizing the quality of the color images constructed from the compressed CFA sampled images. In \sref{sec:results}, we verify the proposed scheme developed in \sref{sec:method} using real raw image sensor data. The preliminary results of this work appeared in \cite{Lee17}.
\section{Background and Review}
\label{sec:background}

\subsection{CFA Sampling}
\label{sec:cfasampling}

\begin{figure}[!t]
\centerline{\includegraphics[scale=1.0]{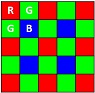}}
\caption{Bayer color filter array pattern~\cite{Bayer75}.}
\label{fig:bayerpattern}
\end{figure}

\begin{figure*}[t]\begin{center}
\begin{minipage}{0.14\linewidth}
  \centering
  \centerline{\includegraphics[scale=1.0]{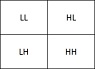}}
  \vspace{1.5cm}
\end{minipage}
\begin{minipage}{0.23\linewidth}
  \centering
  \centerline{\includegraphics[scale=0.9]{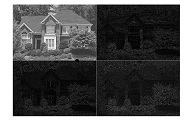}}
\end{minipage}
\begin{minipage}{0.23\linewidth}
  \centering
  \centerline{\includegraphics[scale=0.9]{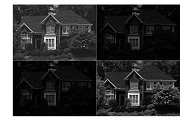}}
\end{minipage}
\begin{minipage}{0.14\linewidth}
  \centering
  \centerline{\includegraphics[scale=1.0]{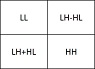}}
  \vspace{1.5cm}
\end{minipage}
\begin{minipage}{0.23\linewidth}
  \centering
  \centerline{\includegraphics[scale=0.9]{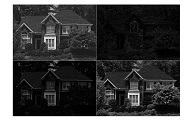}}
\end{minipage}
\\
\begin{minipage}{0.14\linewidth}
  \vspace{3.2cm}
  \centerline{\footnotesize{(a)} }\medskip
\end{minipage}
\begin{minipage}{0.23\linewidth}
  \centering
  \centerline{\includegraphics[scale=0.9]{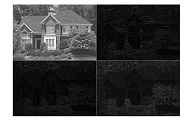}}
  \centerline{\footnotesize{(b) Ordinary wavelet} }\medskip
\end{minipage}
\begin{minipage}{0.23\linewidth}
  \centering
  \centerline{\includegraphics[scale=0.9]{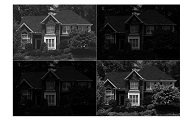}}
  \centerline{\footnotesize{(c)} CFA wavelet}\medskip
\end{minipage}
\begin{minipage}{0.14\linewidth}
  \vspace{3.2cm}
  \centerline{\footnotesize{(d)} }\medskip
\end{minipage}
\begin{minipage}{0.23\linewidth}
  \centering
  \centerline{\includegraphics[scale=0.9]{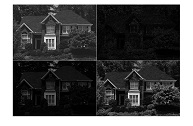}}
  \centerline{\footnotesize{(e) Decorrelated CFA wavelet} }\medskip
\end{minipage}
\caption{The one-level (top row)~LeGall $5/3$ and (bottom row)~Daubechies $9/7$ wavelet transforms. (b) Wavelet coefficients of a typical grayscale image. (c) Wavelet coefficients of a CFA sampled image. (e) Wavelet coefficients of a CFA sampled image with the proposed decorrrelation. The subband layout for (b) and (c) is shown in (a); the layout for (e) is shown in (d). Note that the coefficient magnitudes are scaled for display.}
\vspace{0.2cm} 
\label{fig:DWTofYandCFA}
\end{center}\end{figure*} 

Let $\vec{x}(\vec{n}) = [r(\vec{n})\: g(\vec{n})\: b(\vec{n})]^T$ be a color image at the pixel location $\vec{n} = [n_0 \: n_1]$, where $r(\vec{n})$, $g(\vec{n})$, and $b(\vec{n})$ are the corresponding color tristimulus values of the color. Then the CFA sampled image $y(\vec{n}): \: \mathbb{Z}^2 \rightarrow \mathbb{R}$ is  
\begin{align}\label{eq:cfa1}
y(\vec{n}) = \vec{c}(\vec{n}) \vec{x}(\vec{n}),
\end{align}
where $\vec{c} = [c_r(\vec{n}) \: c_g(\vec{n}) \: c_b(\vec{n})]: \mathbb{Z}^2 \rightarrow \left\{ 0, 1\right\}^3$ is a sampling indicator~(e.g.~$\vec{c}(\vec{n}) = [1 \: 0 \: 0]$ denotes a red sample at a pixel location $\vec{n}$).  

Consider the Bayer pattern shown in \fref{fig:bayerpattern}. The CFA sampled image of \eqref{eq:cfa1} is rewritten as % ~\cite{Dubois05, Korneliussen14} 
\begin{align} \label{eq:cfalumachroma}
\begin{array}{l}
y(\vec{n}) = \vec{c}(\vec{n})
\left[ \begin{array}{ccc}
1 & 2 & 1 \vspace{0.1cm}\\ 1 & 0 & -1 \vspace{0.1cm}\\ 1 & -2 & 1 
\end{array}
\right] \left[ \begin{array}{ccc}
\frac{1}{4} & \frac{1}{2} & \frac{1}{4} \vspace{0.1cm}\\
\frac{1}{4} & 0 & -\frac{1}{4} \vspace{0.1cm}\\
\frac{1}{4} & -\frac{1}{2} & \frac{1}{4}
\end{array} \right] 
\left[ \begin{array}{c}
r(\vec{n}) \vspace{0.1cm}\\ g(\vec{n}) \vspace{0.1cm}\\ b(\vec{n})
\end{array} \right] \vspace{0.2cm}\\ 
\hspace{0.8cm} = \vec{c}'(\vec{n}) \left[ \begin{array}{c}
\ell(\vec{n}) \vspace{0.1cm}\\ \alpha(\vec{n}) \vspace{0.1cm}\\ \beta(\vec{n})
\end{array} \right]. 
\end{array}
\end{align} 
Here, the luminance image $\ell(\vec{n})$ and the color difference images $\alpha(\vec{n})$ and $\beta(\vec{n})$ are defined as 
\begin{align}\label{eq:cfa-analysis}
\begin{array}{l}
\ell(\vec{n}) : = \frac{1}{4} r(\vec{n}) + \frac{1}{2} g(\vec{n}) + \frac{1}{4} b(\vec{n}) \vspace{0.2cm} \\
\alpha(\vec{n}) := \frac{1}{4} r(\vec{n}) - \frac{1}{4} b(\vec{n}) \vspace{0.2cm} \\
\beta(\vec{n}) := \frac{1}{4} r(\vec{n}) - \frac{1}{2} g(\vec{n}) + \frac{1}{4} b(\vec{n}),
\end{array}
\end{align}
where we interpret $\alpha(\vec{n})$, and $\beta(\vec{n})$ as proxies for two chrominance color components. This is convenient way to represent the CFA sampled image because $r(\vec{n})$, $g(\vec{n})$, and $b(\vec{n})$ share high spatial frequency components, yielding lowpass chrominance images $\alpha(\vec{n})$ and $\beta(\vec{n})$~\cite{Gunturk02, Gu10}. The modified CFA mask $\vec{d}(\vec{n})$ is defined as 
\begin{align}\begin{array}{lcl}
\vec{c}'(\vec{n}) & = & \left[ d_\ell \:\: d_\alpha \:\: d_\beta \right] 
: = \vec{c}(\vec{n}) \left[ \begin{array}{ccc} 1 & 2 & 1 \\ 1 & 0 & -1 \\ 1 & -2 & 1 \end{array} \right] \vspace{0.2cm}
\\  
& = & \left[ \begin{array}{ccc} 1 \hspace{0.3cm} (-1)^{n_1}+(-1)^{n_2} \hspace{0.3cm} (-1)^{n_1+n_2} \end{array} \right].
\end{array}
\end{align} 
Hence, the CFA sampled data is rewritten as~\cite{Dubois05, Alleysson05, Korneliussen14}
\begin{align} \label{eq:cfa2}
y(\vec{n}) = \ell(\vec{n}) + d_{\scriptscriptstyle \alpha}(\vec{n})\alpha(\vec{n}) + d_{\scriptscriptstyle \beta}(\vec{n})\beta(\vec{n}). 
\end{align}
In \eqref{eq:cfa2}, the CFA sampled image is interpreted as a linear combination of a complete luminance component and highpass-modulated chrominance components.

\subsection{Wavelet Transform of CFA Sampled Images}
\label{sec:cfadwt}
As observed in \fref{fig:DWTofYandCFA}(c), it is empirically shown that the one-level wavelet transformation of a CFA sampled image is dominated by lowpass signals that are not sparse~\cite{Zhang06}. This observation is contradictory to the wavelet coefficients of ordinary color images in \fref{fig:DWTofYandCFA}(b), where $LH$, $HL$, and $HH$ subbands contain edges and details only. Consequently, the Mallat wavelet packet transform~\cite{Mallat98, Strang96} was proposed to further sparsify the detail coefficients of the CFA sampled images in $HL$, $LH$, and $HH$ subbands. 

A more rigorous analysis of the observation in \cite{Zhang06} was provided in \cite{Hirakawa11}, which we briefly review below. Denote by $w^{y}$ the first level wavelet decomposition of \eqref{eq:cfa2} 
\begin{align}\label{eq:dwtofcfa}
w_{i,j}^{y}(\vec{n}) = w_{i,j}^{\ell}(\vec{n}) + w_{i,j}^{d_{\scriptscriptstyle \alpha} \cdot \alpha}(\vec{n}) + w_{i,j}^{d_{\scriptscriptstyle \beta} \cdot \beta}(\vec{n}), 
\end{align}
where the index $(i,j) \in \{L,H\}^2$ denotes the lowpass~($L$) and highpass~($H$) subbands in the vertical and horizontal directions, and $w^{\ell}$ corresponds to wavelet coefficients of the luminance image $\ell$, etc~\cite{Hirakawa11, Gu10}. The wavelet transform of (\ref{eq:dwtofcfa}) is expanded as the combination of color difference image subbands, as follows:
\begin{align}\label{eq:dwtepn}
\begin{array}{l}
w_{\scriptscriptstyle LL}^{y}(\vec{n}) =  w_{\scriptscriptstyle LL}^{\ell}(\vec{n}) + w_{\scriptscriptstyle LH^\ast}^{\alpha}(\vec{n}) + w_{\scriptscriptstyle H^\ast L}^{\alpha}(\vec{n}) + w_{\scriptscriptstyle H^\ast H^\ast}^{\beta}(\vec{n})\vspace{0.2cm}\\
w_{\scriptscriptstyle LH}^{y}(\vec{n}) =  w_{\scriptscriptstyle LH}^{\ell}(\vec{n}) + w_{\scriptscriptstyle LL^\ast}^{\alpha}(\vec{n}) + w_{\scriptscriptstyle H^\ast H}^{\alpha}(\vec{n}) + w_{\scriptscriptstyle H^\ast L^\ast}^{\beta}(\vec{n}) \vspace{0.2cm} \\
w_{\scriptscriptstyle HL}^{y}(\vec{n}) =  w_{\scriptscriptstyle HL}^{\ell}(\vec{n}) + w_{\scriptscriptstyle L^\ast L}^{\alpha}(\vec{n}) + w_{\scriptscriptstyle HH^\ast}^{\alpha}(\vec{n}) + w_{\scriptscriptstyle L^\ast H^\ast}^{\beta}(\vec{n})\vspace{0.2cm} \\
w_{\scriptscriptstyle HH}^{y}(\vec{n}) =  w_{\scriptscriptstyle HH}^{\ell}(\vec{n}) + w_{\scriptscriptstyle L^\ast H}^{\alpha}(\vec{n}) + w_{\scriptscriptstyle HL^\ast}^{\alpha}(\vec{n}) + w_{\scriptscriptstyle L^\ast L^\ast}^{\beta}(\vec{n}), \vspace{0.2cm} \\
\end{array}
\end{align}
where $L^\ast$ and $H^\ast$ denote the subbands of conjugated wavelet transform coefficients computed using conjugated wavelet filters~\cite{Hirakawa11}. Owing to the lowpass nature of $\alpha$ and $\beta$, the wavelet coefficients of $w^{\alpha}$ and $w^{\beta}$ approximates to zero~($w_{\scriptscriptstyle LH}^{\alpha} \simeq 0, w_{\scriptscriptstyle HL}^{\alpha} \simeq 0$, etc.). Then, the wavelet analysis in \cite{Hirakawa11} yields the following simplification of \eqref{eq:dwtepn}:
\begin{align}\label{eq:dwtsimp}
\begin{array}{l}
w_{\scriptscriptstyle LL}^{y}(\vec{n}) \simeq  w_{\scriptscriptstyle LL}^{\ell}(\vec{n})  \vspace{0.2cm}\\
w_{\scriptscriptstyle LH}^{y}(\vec{n}) \simeq w_{\scriptscriptstyle LH}^{\ell}(\vec{n}) + w_{\scriptscriptstyle LL^\ast}^{\alpha}(\vec{n}) \vspace{0.2cm} \\
w_{\scriptscriptstyle HL}^{y}(\vec{n}) \simeq w_{\scriptscriptstyle HL}^{\ell}(\vec{n}) + w_{\scriptscriptstyle L^\ast L}^{\alpha}(\vec{n}) \vspace{0.2cm} \\
w_{\scriptscriptstyle HH}^{y}(\vec{n}) \simeq w_{\scriptscriptstyle HH}^{\ell}(\vec{n}) + w_{\scriptscriptstyle L^\ast L^\ast}^{\beta}(\vec{n}) \vspace{0.2cm}. \\
\end{array}
\end{align}
Hence, the wavelet coefficients $w_{\scriptscriptstyle LH}^{y}$, $w_{\scriptscriptstyle HL}^{y}$, and $w_{\scriptscriptstyle HH}^{y}$ of a CFA sampled image in \eqref{eq:dwtsimp} are interpreted as sum of fine-scale luminance wavelet coefficients and chrominance conjugate scaling coefficients~\cite{Korneliussen14, Hirakawa07}. Indeed, the wavelet coefficients of $LH$, $HL$, and $HH$ subbands in \fref{fig:DWTofYandCFA}(c) are dominated by lowpass signal of the chrominance images $w_{\scriptscriptstyle LL^\ast}^{\alpha}$ and $w_{\scriptscriptstyle L^\ast L}^{\alpha}$, and $w_{\scriptscriptstyle L^\ast L^\ast}^{\beta}$. 
\subsection{Digital Camera Processing Pipeline}
\label{sec:camera-processing}
Digital camera processing pipeline refers to a sequence of image processing steps that are designed to recover a displayable image from the raw sensor data. We briefly review the basic steps taken, though there are variations among the camera manufacturers on the exact implementations. The steps we describe below are pertinent to the lossy compression algorithm we develop in \sref{sec:pipeline} below.

Let $y(\vec{n})$ denote the raw image sensor data. We first subtract the so-called ``black offset'' $\vec{k}=[k_r \:\: k_g \:\: k_b]^T$, as follows:
\begin{align}
y'(\vec{n})= y(\vec{n})-\vec{c}(\vec{n})\vec{k}.
\end{align}
This step is designed to map the digital values from the analog-to-digital converter (ADC) to a number that is linear to the light intensity so that a $y'(\vec{n})$ would map to zero when there is no light coming into the camera. 

Following the black offset, the step known as ``demosaicking'' estimates the color image $\vec{x}(\vec{n})$ from the CFA sampled image $y'(\vec{n})$~\cite{Korneliussen14, Hirakawa07, Alleysson05, Gunturk02, Dubois05, Gu10}. In a step known as ``color correction'', the colors of the recovered image $\widehat{\vec{x}}(\vec{n})$ corresponding to the spectral transmittance of the color filters are converted to a canonical color space by multiplying by a color transformation matrix $\mat{A} \in \mathbb{R}^{3 \times 3}$:
\begin{align}
\vec{x}_{cc}(\vec{n}) = \mat{A}\widehat{\vec{x}}(\vec{n}).
\end{align}
The ``white balance'' step rescales the color to make the color (nearly) invariant to illumination color~\cite{Choudhury10, Finlayson93, Land83}, 
\begin{align}
\vec{x}_{wb,cc}(\vec{n}) = 
\begin{bmatrix}
1/i_r&0&0\\
0&1/i_g&0\\
0&0&1/i_b
\end{bmatrix}
\vec{x}_{cc}(\vec{n}).
\end{align}
where $[i_r,i_g,i_b]$  is the color of the illumination. Lastly, a compander known as gamma correction enhances the low-intensity pixels while compressing the high-intensity pixels by a non-linear mapping~\cite{Dufaux16}. For example, gamma correction used in sRGB standard take the form~\cite{sRGB}:
\begin{align}
\vec{x}_{gc,wb,cc}(\vec{n}) = \left\{ 
\begin{array}{l}
12.92\vec{x}_{wb,cc}(\vec{n}), \:\: \textrm{if} \:\:\vec{x}_{wb,cc}(\vec{n}) \leq 0.0031308  \vspace{0.2cm} \\
(1+0.055)\vec{x}_{wb,cc}(\vec{n})^{1/2.4} - 0.055, \:\: \textrm{otherwirse}.
\end{array}
\right.
\end{align}
Although other steps (e.g.~defective pixel removal and noise suppression) are also common, black offset, demosaicking, color correction, white balance, and gamma correction are considered bare minimum steps. The processed color image $\vec{x}_{gc,wb,cc}(\vec{n})$ is considered the final output of the digital camera---it is either displayed or compressed and stored.
\section{Proposed Compression Scheme of CFA Sampled Image}
\label{sec:method}
The wavelet analysis in \sref{sec:cfasampling} and \sref{sec:cfadwt} suggests that the wavelet coefficients of a CFA sampled image are combinations of their luminance wavelet component and modulated frequency chrominance component. This implies that the $LH$, $HL$, and $HH$ subbands directly yield poor coding efficiency, because $\omega_{\scriptscriptstyle LL^\ast}^{\alpha}$ and $\omega_{\scriptscriptstyle L^\ast L}^{\alpha}$ would never achieve the coding efficiency of $\omega_{LH}^{g}$ and $\omega_{HL}^{g}$ even if further wavelet transform is applied. Hence, we instead propose to decorrelate $LH$ and $HL$ subbands based on the wavelet analysis in Section~\ref{sec:background}. 

In \sref{sec:lossless}, we develop a lossless compression method based on the decorrelated Mallat wavelet structure. We extend this idea to lossy compression in \sref{sec:lossy}, where we achieve superior PSNR on the reconstruction of the CFA sampled image. In \sref{sec:alternative}, we detail implementational optimizations. In \sref{sec:pipeline}, we further optimize the lossy compression minimizing the error of the reconstructed color image \emph{after} it has undergone the camera processing pipeline.

\begin{figure*}[t]
\begin{minipage}{0.24\linewidth}
  \centering
  \centerline{\includegraphics[scale=0.2]{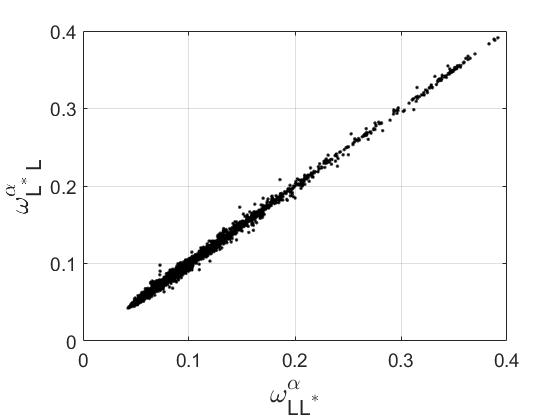}}
  \centerline{\footnotesize{(a)} }\medskip
  % 0.9989
\end{minipage}
\begin{minipage}{0.24\linewidth}
  \centering
  \centerline{\includegraphics[scale=0.2]{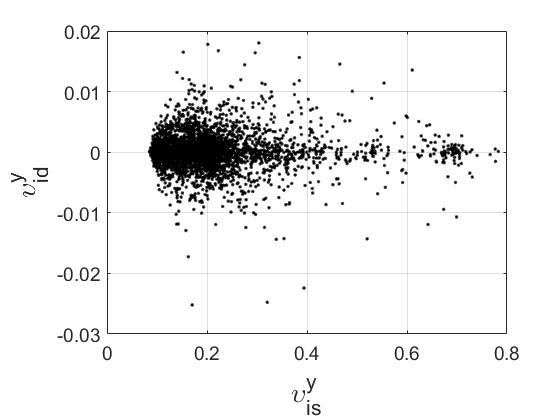}}
  \centerline{\footnotesize{(b)} }\medskip
  % -0.0174
\end{minipage}
\begin{minipage}{0.24\linewidth}
  \centering
  \centerline{\includegraphics[scale=0.2]{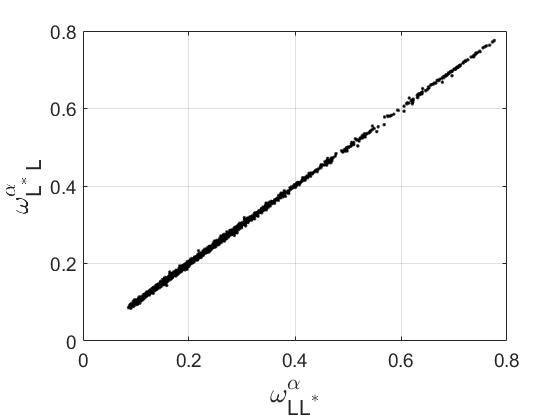}}
  \centerline{\footnotesize{(c)} }\medskip
  % 0.9998
\end{minipage}
\begin{minipage}{0.24\linewidth}
  \centering
  \centerline{\includegraphics[scale=0.2]{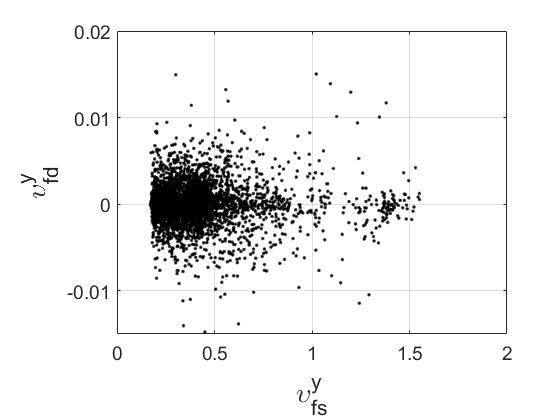}}
  \centerline{\footnotesize{(d)} }\medskip
  % 0.0141
\end{minipage}
\\
\begin{minipage}{0.24\linewidth}
  \centering
  \centerline{\includegraphics[scale=0.2]{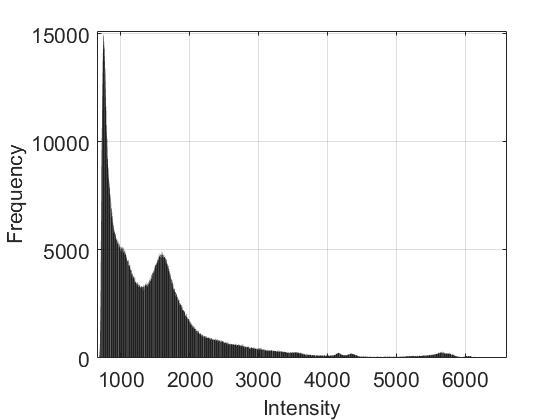}}
  \centerline{\footnotesize{(e)} }\medskip
  % 0.9989
\end{minipage}
\begin{minipage}{0.24\linewidth}
  \centering
  \centerline{\includegraphics[scale=0.2]{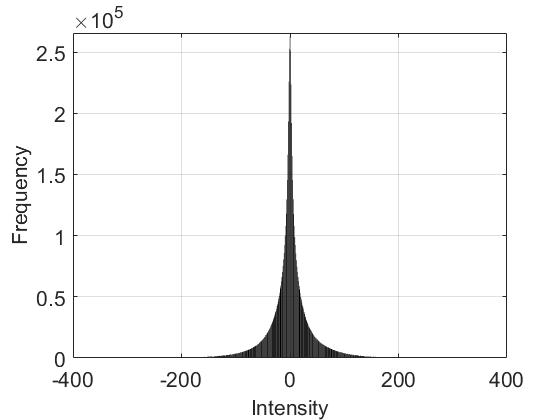}}
  \centerline{\footnotesize{(f)} }\medskip
  % -0.0174
\end{minipage}
\begin{minipage}{0.24\linewidth}
  \centering
  \centerline{\includegraphics[scale=0.2]{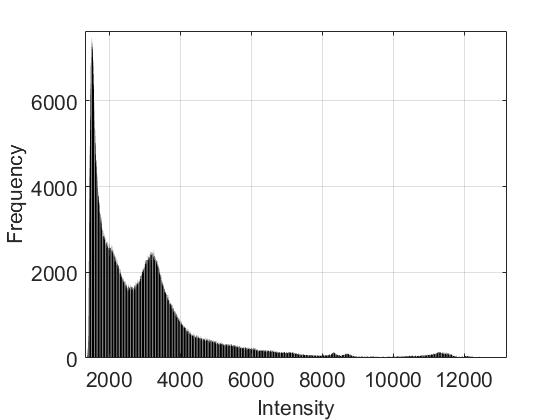}}
  \centerline{\footnotesize{(g)} }\medskip
  % 0.9998
\end{minipage}
\begin{minipage}{0.24\linewidth}
  \centering
  \centerline{\includegraphics[scale=0.2]{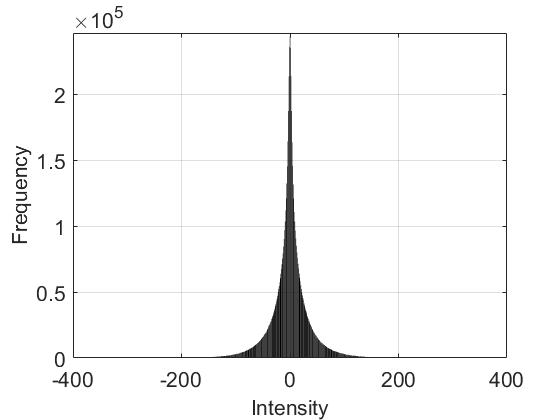}}
  \centerline{\footnotesize{(h)} }\medskip
  % 0.0141
\end{minipage}

\caption{Illustration of high redundancy in the Mallat wavelet decomposition. (a) Plot of $w_{\scriptscriptstyle LL^\ast}^{\alpha}$ against $w_{\scriptscriptstyle L^\ast L}^{\alpha}$ for the LeGall 5/3 wavelet transform. (b) Plot of $v_{\scriptscriptstyle s}^{y}$ against $v_{\scriptscriptstyle d}^{y}$ for the LeGall 5/3 wavelet transform. (c) Plot of $w_{\scriptscriptstyle LL^\ast}^{\alpha}$ against $w_{\scriptscriptstyle L^\ast L}^{\alpha}$ for the Daubechies 9/7 wavelet transform. (d) Plot of $v_{\scriptscriptstyle s}^{y}$ against $v_{\scriptscriptstyle d}^{y}$ for the Daubechies 9/7 wavelet transform. (e) Histogram of $w_{\scriptscriptstyle LL^\ast}$ for the LeGall 5/3 wavelet transform. (f) Histogram of $w_{\scriptscriptstyle LL^\ast}^{\alpha} - w_{\scriptscriptstyle L^\ast L}^{\alpha}$ for the LeGall 5/3 wavelet transform. (g) Histogram of $w_{\scriptscriptstyle LL^\ast}$ for the Daubechies 9/7 wavelet transform. (h) Histogram of $w_{\scriptscriptstyle LL^\ast}^{\alpha} - w_{\scriptscriptstyle L^\ast L}^{\alpha}$ for the Daubechies 9/7 wavelet transform. }
\label{fig:correaltion}
\end{figure*}
\begin{figure*}[!ht]
\begin{minipage}{0.24\linewidth}
  \centering
  \centerline{\includegraphics[scale=0.2]{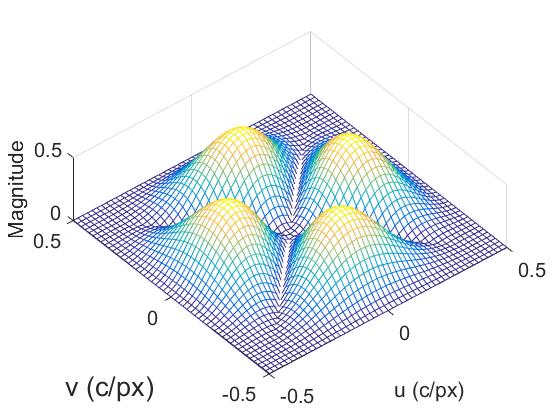}}
  \vspace{0.2cm}
  \centerline{\footnotesize{(a) $LL^\ast - L^\ast L$} }\medskip
  \vspace{0.2cm}
\end{minipage}
\begin{minipage}{0.24\linewidth}
  \centering
  \centerline{\includegraphics[scale=0.2]{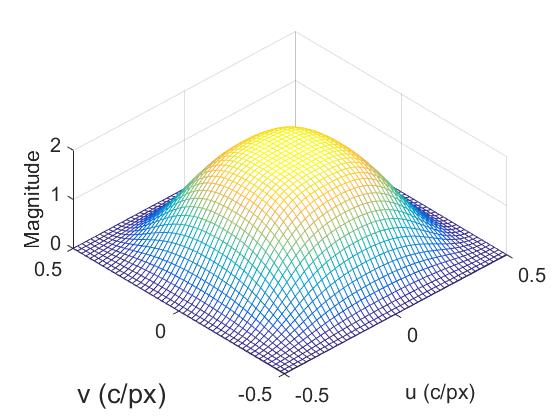}}
  \vspace{0.2cm}
  \centerline{\footnotesize{(b) $LL^\ast + L^\ast L$} }\medskip
  \vspace{0.2cm}
\end{minipage}
\begin{minipage}{0.24\linewidth}
  \centering
  \centerline{\includegraphics[scale=0.2]{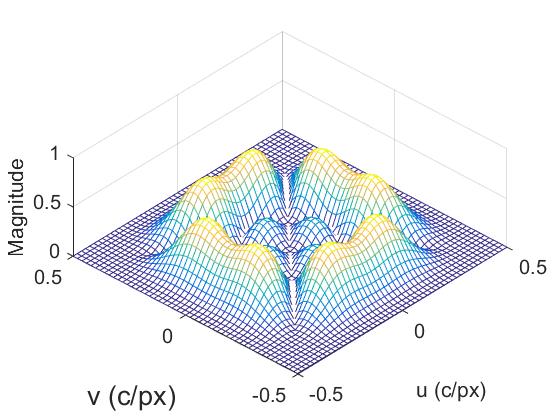}}
  \vspace{0.2cm}
  \centerline{\footnotesize{(c) $LL^\ast - L^\ast L$} }\medskip
  \vspace{0.2cm}
\end{minipage}
\begin{minipage}{0.24\linewidth}
  \centering
  \centerline{\includegraphics[scale=0.2]{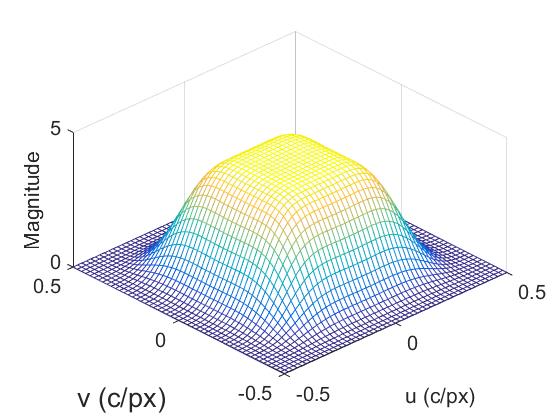}}
  \vspace{0.2cm}
  \centerline{\footnotesize{(d) $LL^\ast + L^\ast L$} }\medskip
  \vspace{0.2cm}
\end{minipage}
\\
\begin{minipage}{0.24\linewidth}
  \centering
  \centerline{\includegraphics[scale=1.0]{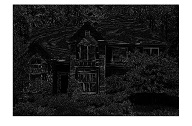}}
  \centerline{\footnotesize{(e) $w_{\scriptscriptstyle LL^\ast}^{\alpha} - w_{\scriptscriptstyle L^\ast L}^{\alpha}$} }\medskip
\end{minipage}
\begin{minipage}{0.24\linewidth}
  \centering
  \centerline{\includegraphics[scale=1.0]{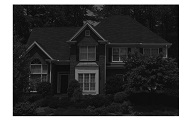}}
  \centerline{\footnotesize{(f) $w_{\scriptscriptstyle LL^\ast}^{\alpha} + w_{\scriptscriptstyle L^\ast L}^{\alpha}$} }\medskip
\end{minipage}
\begin{minipage}{0.24\linewidth}
  \centering
  \centerline{\includegraphics[scale=1.0]{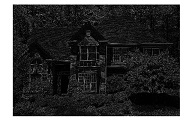}}
  \centerline{\footnotesize{(g) $w_{\scriptscriptstyle LL^\ast}^{\alpha} - w_{\scriptscriptstyle L^\ast L}^{\alpha}$} }\medskip
\end{minipage}
\begin{minipage}{0.24\linewidth}
  \centering
  \centerline{\includegraphics[scale=1.0]{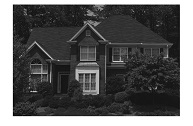}}
  \centerline{\footnotesize{(h) $w_{\scriptscriptstyle LL^\ast}^{\alpha} + w_{\scriptscriptstyle L^\ast L}^{\alpha}$} }\medskip
\end{minipage}
\caption{Filter responses and examples of the filtered coefficients.  The spatial frequencies $u$ and $v$ for respectively $n_0$ and $n_1$ are normalized in the range $-1.0$ to $1.0$, where $1.0$ corresponds to $\pi$ radians in cycles per pixel height~(c/px~\cite{Dubois05}). (a), (b), (e), (f) The LeGall 5/3 wavelet transform.  (c), (d), (g), (h) The Daubechies 9/7 wavelet transform. The intensity of the filtered coefficients are scaled for display. }
\label{fig:residualerror}
\end{figure*} 

\subsection{Lossless Compression of CFA sampled images}
\label{sec:lossless}
Recall that $w_{\scriptscriptstyle LL^\ast}^{\alpha}$ and $w_{\scriptscriptstyle L^\ast L}^{\alpha}$ are scaling coefficients of the same chrominance image $\alpha$ using two conjugate variations of the wavelet transforms. This suggests that the $w_{\scriptscriptstyle LL^\ast}^{\alpha}$ and $w_{\scriptscriptstyle L^\ast L}^{\alpha}$ are highly correlated, which may decrease coding efficiency. Examining \eqref{eq:dwtsimp} using the LeGall $5/3$ wavelet transforms, the plot in \fref{fig:correaltion}(a) demonstrate that $w_{\scriptscriptstyle LL^\ast}^{\alpha}$ in $w_{\scriptscriptstyle LH}^{y}$ and $w_{\scriptscriptstyle L^\ast L}^{\alpha}$ in $w_{\scriptscriptstyle HL}^{y}$ are highly correlated. Indeed, the strength of correlation by the Pearson product-moment correlation coefficients is $0.9989$~\cite{Stark11, Garcia08}. 

In lossless compression, the coefficients of $w_{\scriptscriptstyle HL}^{y}$ and $w_{\scriptscriptstyle LH}^{y}$ are decorrelated by orthogonal transformation using the sums and differences. To sparsify $w_{\scriptscriptstyle LH}^{y}$ and $w_{\scriptscriptstyle HL}^{y}$, they are replaced by the decorrelated coefficients $v_{\scriptscriptstyle d}^{y}$ and $v_{\scriptscriptstyle s}^{y}$, as follows: 
\begin{align} \label{eq:deco-lossless}
\begin{array}{l}
v_{\scriptscriptstyle d}^{y}(\vec{n}) = w_{\scriptscriptstyle LH}^{y}(\vec{n}) - w_{\scriptscriptstyle HL}^{y}(\vec{n}) \vspace{.1cm} \\ 
\hspace{1.0 cm} =  \Big( w_{\scriptscriptstyle LH}^{\ell}(\vec{n}) - w_{\scriptscriptstyle HL}^{\ell}(\vec{n})\Big) + \Big( w_{\scriptscriptstyle LL^\ast}^{\alpha}(\vec{n})- w_{\scriptscriptstyle L^\ast L}^{\alpha}(\vec{n}) \Big)\vspace{0.2cm} \\
v_{\scriptscriptstyle s}^{y}(\vec{n}) = \Big\lfloor \frac{1}{2} \left( w_{\scriptscriptstyle LH}^{y}(\vec{n}) + w_{\scriptscriptstyle HL}^{y}(\vec{n})\right)  \Big\rfloor  \vspace{0.2cm} \\
\hspace{0.5cm} = \Bigl\lfloor \frac{1}{2} \Big( w_{\scriptscriptstyle LH}^{\ell}(\vec{n}) + w_{\scriptscriptstyle HL}^{\ell}(\vec{n}) \Big) + \frac{1}{2} \Big( w_{\scriptscriptstyle LL^\ast}^{\alpha}(\vec{n}) + w_{\scriptscriptstyle L^\ast L}^{\alpha}(\vec{n}) \Big) \Bigl\rfloor, \vspace{0.2cm} \\
\end{array}
\end{align} 
where $\lfloor \cdot \rfloor$ denotes a $floor$ operation. Then, $w_{\scriptscriptstyle HL}^{y}$ and $w_{\scriptscriptstyle LH}^{y}$ are perfectly reconstructable from the decorrelated coefficients $v_{\scriptscriptstyle d}^{y}$ and $v_{\scriptscriptstyle s}^{y}$ as 
\begin{align}
\begin{array}{lcl}
w_{\scriptscriptstyle HL}^{y}(\vec{n}) & = & v_{s}^{y}(\vec{n}) - \left\lfloor \frac{v_{d}^{y}(\vec{n})}{2} \right\rfloor, \vspace{0.2cm} \\
w_{\scriptscriptstyle LH}^{y}(\vec{n}) & = & v_{d}^{y}(\vec{n}) + w_{\scriptscriptstyle HL}^{y}(\vec{n}) . 
\end{array}   
\end{align}

\begin{figure*}[!t]
\centerline{\includegraphics[scale=0.4]{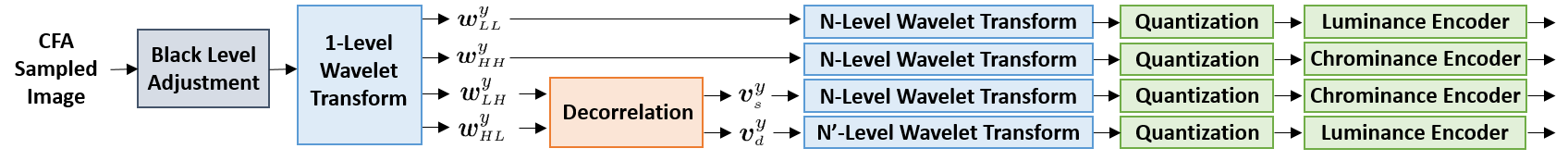}}
\caption{Description of the proposed compression schemes in Sections~\ref{sec:lossless} and \ref{sec:lossy}. In lossless compression, the quantization step is not performed.}
\label{fig:schematicreps}
\end{figure*}

% difference subband
The difference subbands $v_{\scriptscriptstyle d}^{y}$ is designed to decorrelate $w_{\scriptscriptstyle LL^\ast}^{\alpha}$ and $w_{\scriptscriptstyle L^\ast L}^{\alpha}$. This is confirmed by \fref{fig:correaltion}(b), and the Pearson product-moment correlation coefficient decreased to $-0.0174$. The coding efficiency also increases as demonstrated by the entropy of $6.99$ in \fref{fig:correaltion}(f) relative to $11.05$ of \fref{fig:correaltion}(e). To understand why this is the case, consider rewriting the difference $w_{\scriptscriptstyle LL^\ast}^{\alpha} - w_{\scriptscriptstyle L^\ast L}^{\alpha}$ as the output of a single filtering operation on the signal $\alpha$ by a difference of lowpass filters $LL^\ast - L^\ast L$. As evidenced by the frequency responses shown in \fref{fig:residualerror}(a), the difference of lowpass filters forms a bandpass filter, hence suppressing the lowpass signal $\alpha$. We conclude that the residual $w_{\scriptscriptstyle LL^\ast}^{\alpha} - w_{\scriptscriptstyle L^\ast L}^{\alpha}$ in $v_{\scriptscriptstyle d}^{y}$ is small~(see \fref{fig:residualerror}(e)). Appealing to the fact that $v_{\scriptscriptstyle d}^{y}$ is dominated by the luminance highpass $w^\ell_{\scriptscriptstyle LH}$ and $w^\ell_{\scriptscriptstyle HL}$ with a very limited contribution from the bandpass chrominance $w^\alpha_{\scriptscriptstyle LL^\ast}-w^\alpha_{\scriptscriptstyle L^\ast L}$, it requires only minimal additional levels ($N'$ in \fref{fig:schematicreps}) of wavelet transforms (LeGall 5/3 in our implementation) to sparsify $v_d^y$. Therefore, we propose to encode $v_{\scriptscriptstyle d}^{y}$ the same way we would normally encode $w^\ell_{\scriptscriptstyle LH}$ or $w^\ell_{\scriptscriptstyle HL}$. See \fref{fig:schematicreps}.

% sum subband
On the other hand, the sum subband $v_{\scriptscriptstyle s}^{y}$ is comprised of the lowpass chrominance $\alpha$ and the highpass luminance $\ell$. The sum $w^\alpha_{\scriptscriptstyle LL^\ast}+w^\alpha_{\scriptscriptstyle L^\ast L}$ can be thought of as the filtering output of the chrominance signal $\alpha$ through a single filter $LL^\ast+L^\ast L$. As shown by the frequency responses in \fref{fig:residualerror}(b) and the filtered images in \fref{fig:residualerror}(f), the sum of lowpass filters $LL^\ast + L^\ast L$ also corresponds to a lowpass filter. Hence, the wavelet coefficient $v_{\scriptscriptstyle s}^{y}$ is a mixture of lowpass chrominance and highpass luminance. We argue that compression should treat $v_{s}^{y}$ as we treat the chrominance of $w^\alpha_{\scriptscriptstyle LL}$ because the relative magnitudes of the fine scale wavelet coefficients $w^\ell_{\scriptscriptstyle LH}$ and $w^\ell_{\scriptscriptstyle HL}$ are far smaller than the chrominance $w_{\scriptscriptstyle LL}^{\alpha}$. Hence, we proposed to apply additional N levels of wavelet transforms to
$v_{s}^{y}$~(where $N>N'$), in a manner similar to the Mallat wavelet packet transform~\cite{Zhang06}. The Mallat wavelet packet transform coefficients are subsequently encoded by a variable length coding scheme designed for a chrominance component~(similar to the way $w_{\scriptscriptstyle LL}^{\alpha}$ would normally be treated). See \fref{fig:schematicreps}.   

% LL and HH subband
Finally, the wavelet coefficients $w_{\scriptscriptstyle LL}^{y}$ and $w_{\scriptscriptstyle HH}^{y}$ of (\ref{eq:dwtsimp}) are interpreted as low frequency components of luminance $w_{\scriptscriptstyle LL}^{\ell}$ and chrominance $w_{\scriptscriptstyle L^\ast L^\ast}^{\beta}$, respectively. Therefore, additional $N$ levels of wavelet transform should be applied to $w_{\scriptscriptstyle LL}^{y}$ and $w_{\scriptscriptstyle HH}^{y}$, similar in manner to \cite{Zhang06}. We apply a encoding scheme of luminance and chrominance to them, respectively. See \fref{fig:schematicreps}.

% black offset
In our compression method, we shift each color component of the CFA sampled image by adjusting its offset before taking wavelet transform, as follows:
\begin{align}\label{eq:blackoffset}
y'(\vec{n}) = y(\vec{n}) -\vec{c}(\vec{n})\vec{k}
\end{align}
where $\vec{k} = \left[ k_r \: k_g \: k_b \right]^{T}$ is the integer offset values of color components~(this is commonly performed in camera processing pipelines already). The shift $\vec{k}$ must be stored as a sideband information in order to uncompress the image later. In the experiment, the values of $\vec{k}$ are chosen by black offset, which was computed from a calibration experiment using X-Rite ColorChecker\footnote{\url{www.xritephoto.com}}. This has the effect of shifting $v_{\scriptscriptstyle s}^{y} \simeq \left(  w_{\scriptscriptstyle LL^\ast}^{\alpha}+w_{\scriptscriptstyle L^\ast L}^{\alpha}\right)/2$ and $w_{\scriptscriptstyle HH}^{y} \simeq w_{\scriptscriptstyle L^\ast L^\ast}^{\beta}$ toward zero, which increases the coding efficiency further. See \fref{fig:schematicreps}. 

The image is uncompressed by following the steps in \fref{fig:schematicreps} in the reverse order.

\subsection{Lossy Compression of CFA sampled images}
\label{sec:lossy}
The proposed compression scheme in lossy mode is identical to the lossless scheme in \sref{sec:lossless} as illustrated in \fref{fig:schematicreps}, except for the decorrelation step and the quantization step. The wavelet coefficients $w_{\scriptscriptstyle LL^\ast}^{\alpha}$ in $w_{\scriptscriptstyle LH}^{y}$ and $w_{\scriptscriptstyle L^\ast L}^{\alpha}$ in $w_{\scriptscriptstyle HL}^{y}$ of the Daubechies 9/7 wavelet transform are also highly correlated, as demonstrated by the strength of the Pearson product-moment correlation coefficient of $0.9989$ in \fref{fig:correaltion}(c). 

In lossy compression, the $w_{\scriptscriptstyle LH}^{y}$ and $w_{\scriptscriptstyle HL}^{y}$ are decorrelated by a non-integer linear transformation. That is, let 
\begin{align}
\left[ \begin{array}{c} v_{\scriptscriptstyle s}^{y}(\vec{n}) \vspace{0.1cm}\\ v_{\scriptscriptstyle d}^{y}(\vec{n})  \end{array} \right]
= \mat{M}
\left[ \begin{array}{c} w_{\scriptscriptstyle LH}^{y}(\vec{n}) \vspace{0.1cm}\\ w_{\scriptscriptstyle HL}^{y}(\vec{n})  \end{array} \right], 
\end{align} 
where $\mat{M} \in \mathbb{R}^{2 \times 2}$. Recalling that we quantize $v_{\scriptscriptstyle s}^{y}$ and $v_{\scriptscriptstyle d}^{y}$~(not $w_{\scriptscriptstyle LH}^{y}$ and $w_{\scriptscriptstyle HL}^{y}$), let $\hat{v}_{\scriptscriptstyle s}^{y}(\vec{n}) = v_{\scriptscriptstyle s}^{y}(\vec{n}) + q_s(\vec{n})$ and $\hat{v}_{\scriptscriptstyle d}^{y}(\vec{n}) = v_{\scriptscriptstyle d}^{y}(\vec{n}) + q_d(\vec{n})$ denote the quantized versions of $v_{\scriptscriptstyle s}^{y}$ and $v_{\scriptscriptstyle d}^{y}$, respectively, where $q_{\scriptscriptstyle s}$ and $q_{\scriptscriptstyle d}$ are the quantization errors approximated as uniform distribution. By the choice of $\mat{M}$, we minimize the error between the wavelet coefficients $w$ and its quantized coefficients while penalizing non-sparsity of the decorrelated coefficients $v$, as follows: 
\begin{align}\label{eq:select-m}
\begin{array}{l}
\underset{\mat{M}}{\arg\min} \:\: \sum\limits_{n} \Bigg|\Bigg|
\left[\!\! \begin{array}{c} w_{\scriptscriptstyle LH}^{y}(\vec{n}) \vspace{0.1cm}\\ w_{\scriptscriptstyle HL}^{y}(\vec{n}) \end{array} \!\! \right]
\!\! - \!\! \mat{M}^{-1} \!\! 
\left[\!\! \begin{array}{c} v_{\scriptscriptstyle s}^{y}(\vec{n}) + q_{\scriptscriptstyle s}(\vec{n}) \vspace{0.1cm} \\ 
v_{\scriptscriptstyle d}^{y}(\vec{n}) + q_{\scriptscriptstyle d}(\vec{n}) \end{array} \!\! \right]
\!\! \Bigg|\Bigg|_{2}^{2} \vspace{0.1cm} \\ 
\hspace{2cm} + \lambda \sum\limits_{n} \Bigg|\Bigg| \left[\!\! \begin{array}{c} v_{\scriptscriptstyle s}^{y}(\vec{n}) \vspace{0.1cm}\\ v_{\scriptscriptstyle d}^{y}(\vec{n}) \end{array} \!\! \right] \Bigg|\Bigg|_{1} \vspace{0.2cm}\\
\simeq \underset{\mat{M}}{\arg\min} \:\: \mathbb{E} \Bigg\{ \sum\limits_{n}\Bigg(\, \Bigg|\Bigg| 
\mat{M}^{-1} \!\! 
\left[\!\! \begin{array}{c} q_{\scriptscriptstyle s}(\vec{n}) \vspace{0.1cm} \\ 
q_{\scriptscriptstyle d}(\vec{n}) \end{array} \!\! \right]
\!\! \Bigg|\Bigg|_{2}^{2} + \lambda \Bigg|\Bigg| \left[\!\! \begin{array}{c} v_{\scriptscriptstyle s}^{y}(\vec{n}) \vspace{0.1cm}\\ v_{\scriptscriptstyle d}^{y}(\vec{n}) \end{array} \!\! \right] \Bigg|\Bigg|_{1}  \,\Bigg) \Bigg\}  \vspace{0.3cm}\\
= \underset{\mat{M}}{\arg\min} \:\: \big|\big| \mat{M} \big|\big|_{F}^{2} 
+ \lambda \sum\limits_{n} \Bigg|\Bigg| \left[ \begin{array}{c} v_{\scriptscriptstyle s}^{y}(\vec{n}) \vspace{0.1cm}\\ v_{\scriptscriptstyle d}^{y}(\vec{n}) \end{array} \right] \Bigg|\Bigg|_{1}.
\end{array}
\end{align}
Here, the approximation stems from the law of large numbers~(as denoted by the expectation operator $\mathbb{E}\{\cdot\}$)~\cite{Stark11, Garcia08} applied to the quantization errors $q_{\scriptscriptstyle s}$ and $q_{\scriptscriptstyle d}$. The simplification by the Frobenius norm $||\cdot ||_F^2$ stems from the assumption that $q_s$ and $q_d$ are uniformly distributed and independent. Increasing the value of $\lambda$ promotes sparsity~(and coding efficiency) at the sacrifice of the reconstruction error. In practice, we found that $\mat{M} \simeq k \cdot \left[ \begin{array}{cc} a & a \vspace{0.1cm}\\ b & -b
\end{array} \right]$, where the transformation of $a$ and $b$ were stable while $k$ decreased with increasing $\lambda$, where $v_s^y\approx ka(w_{\scriptscriptstyle LH}^{\alpha}+w_{\scriptscriptstyle HL}^{\alpha})$ and $v_d^y\approx kb(w_{\scriptscriptstyle LH}^\ell-w_{\scriptscriptstyle HL}^\ell)$. We also found that the minimization in \eqref{eq:select-m} can be performed numerically by gradient descent. 

The transformation $\mat{M}$ decorrelates the coefficients $v_{s}^{y}$ and $v_{d}^{y}$ as evidenced by the plot of \fref{fig:correaltion}(d) and the Pearson product-moment correlation coefficient decreased to $0.014$. The entropy of the decorrelated coefficient reduced to $6.91$ of \fref{fig:correaltion}(h) relative to $12.05$ of \fref{fig:correaltion}(g). 

\begin{figure*}[!t]
\centerline{\includegraphics[scale=0.4]{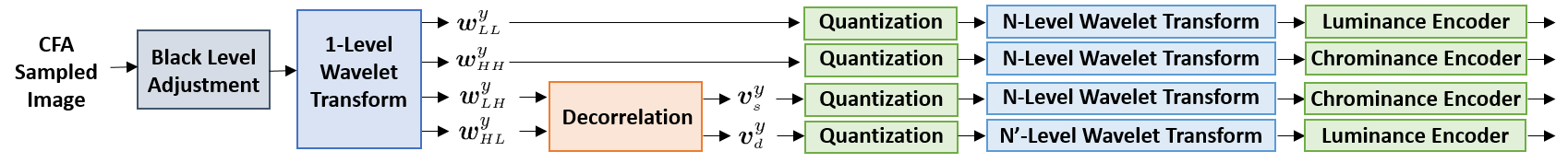}}
\caption{Description of the ad-hoc alternative lossy compression scheme.}
\label{fig:ad-hoc-pipeline}
\end{figure*}

\subsection{Optimization of Wavelet Transforms}
\label{sec:alternative}
There are two main sources of distortions in lossy compression: round-off error and quantization error. The round-off error stems from finite precision operators used to carry out the forward and reverse wavelet transforms. The quantization error~(commonly referred to as the ``residual error'') is caused by reducing the number of bits to represent wavelet coefficients, at the expense of accuracy. Specifically, a larger quantization step yields higher compression ratio and higher loss in quality. 

The interactions between the two sources of noise depend on the bitrate. Although the quantization errors dominate at the low bitrates, the round-off error limits the image quality at the higher bitrates. This suggests that better quality would be achieved if the round-off error is reduced at the higher bitrates. By experimentation, we heuristically arrived at an alternative decomposition scheme as illustrated in \fref{fig:ad-hoc-pipeline} that perform better at high bitrates. We propose quantizing the coefficients after the first level Daubechies $9/7$ wavelet transform and decorrelation step outlined in \sref{sec:lossy}. We then use the LeGall $5/3$ to carry out the Mallat wavelet packet transform on $w_{\scriptscriptstyle LL}^y$, $w_{\scriptscriptstyle HH}^y$, and $v_s^y$. Since LeGall $5/3$ transform is reversible, no additional round-off errors are expected~(or post-pipeline quarter resolution color image, as described in \sref{sec:pipeline}). This is in contrast to the conventional compression scheme of quantizing the coefficients after multiple-level wavelet transform~\cite{Malvar12}. 

As the quantization step increases, the round-off error become insignificant relative to the quantization error. Hence at the lower bitrates, we empirically found that Daubechies $9/7$ would be more effective for the decorrelated Mallat as described in \sref{sec:lossy} and in \fref{fig:schematicreps}. The trade-offs between bitrates and the choice of wavelet transforms are confirmed by our experiments in \sref{sec:lossy-results}. 
\subsection{Camera Processing Pipeline-Aware Lossy Compression}
\label{sec:pipeline}
\begin{figure*}[!t]
\centerline{\includegraphics[scale=0.4]{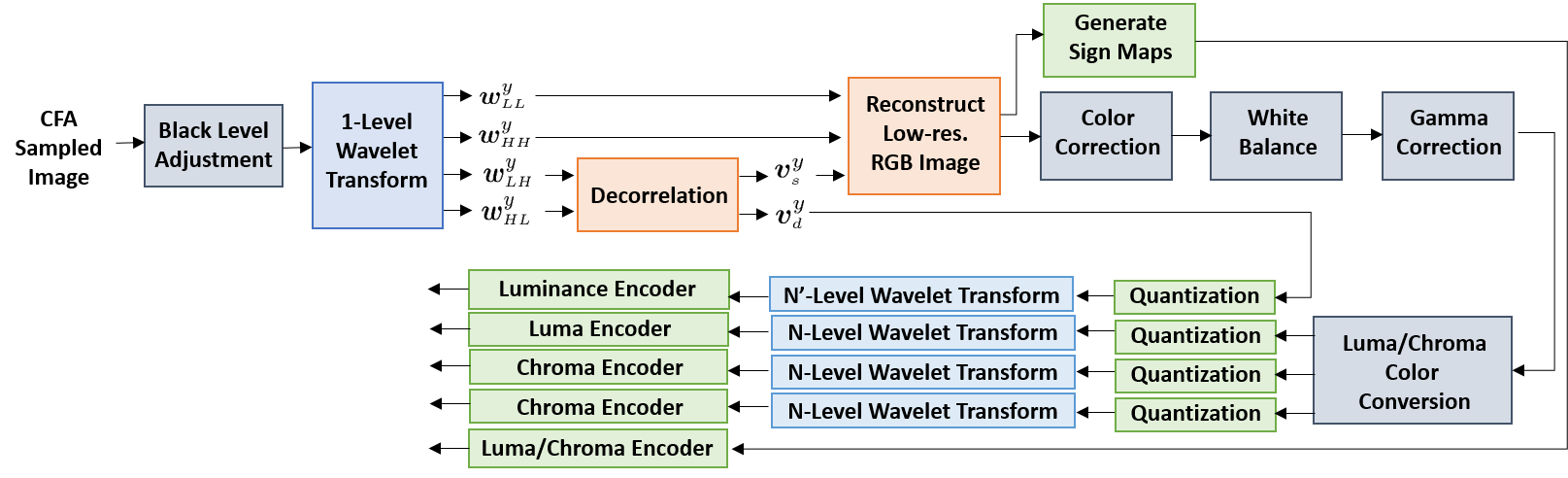}}
\caption{Description of the proposed compression scheme of CAMRA in Section~\ref{sec:pipeline}.}
\label{fig:cfa-pipeline}
\end{figure*}
\begin{figure}[t]
\begin{minipage}{0.6\linewidth}
\includegraphics[scale=0.45]{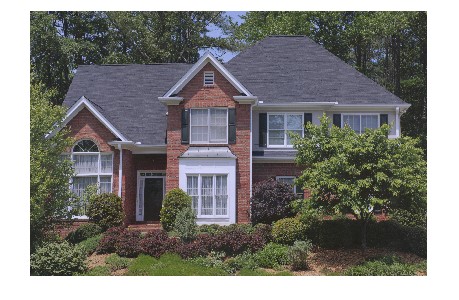}\\
\end{minipage}
\begin{minipage}{0.3\linewidth}
\includegraphics[scale=0.45]{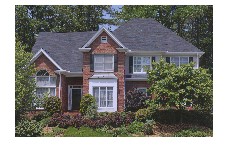}\\
\vspace{1.6cm}
\end{minipage}
\vspace{0.1cm}
\caption{Comparison of the color image at the end of the pipeline in \sref{sec:camera-processing}~(left) and the quarter resolution color image recovered from the wavelet coefficients in \sref{sec:pipeline}~(right).}
\label{fig:rgb-quarter-rgb}
\end{figure}

The decorrelated decomposition schemes in Sections~\ref{sec:lossless} and \ref{sec:lossy} improve coding efficiency relative to the existing CFA compression schemes. Recalling that the color image is constructed using a decompressed CFA sampled image, there may be additional penalties in image quality when the distortion of the lossy compression is propagated through the camera processing pipeline.  Specifically, CFA sampled images are not invariant to the color of illumination, raising the possibility that one color component is potentially compressed more than other components. Similarly, color correction would boost errors in a certain color channel relative to others. The human eyes are more sensitive to the dark regions of the image, which is further exaggerated by gamma correction. Clearly, uniformly quantizing the decorrelated Mallat wavelet packet transform coefficients of the CFA sampled image is suboptimal for minimizing the distortion of the color image we recover from the (decompressed) raw sensor data.

Hence, in this section, we develop a pipeline for compressing CFA sampled images that maximizes the quality of the subsequently reconstructed color images, based on {\it camera-aware multi-resolution analysis~(CAMRA)}. The schematic representation of the proposed pipeline is shown in \fref{fig:cfa-pipeline}. Recalling \eqref{eq:cfa-analysis}, the color components $\vec{x}(\vec{n})$ are reconstructable from $\ell$, $\alpha$, and $\beta$ by the relation:
\begin{align} \label{eq:reverse-rgb}
\vec{x}(\vec{n}) = \left[ \begin{array}{ccc} 1 & 2 & 1\vspace{0.1cm}\\ 1 & 0 & -1\vspace{0.1cm}\\ 1 & -2 & 1 \end{array} \right]
\left[ \begin{array}{c} \ell(\vec{n}) \vspace{0.1cm}\\ \alpha(\vec{n})\vspace{0.1cm}\\ \beta(\vec{n}) \end{array} \right].
\end{align}
Suppose we interpret the $w_{\scriptscriptstyle LL}^y\approx w_{\scriptscriptstyle LL}^\ell$, $w_{\scriptscriptstyle HH}^y\approx w_{\scriptscriptstyle LL}^\beta,$ and $v_s^y\approx w_{\scriptscriptstyle LL}^\alpha$ (or $2ka w_{\scriptscriptstyle LL}^\alpha$ if lossy) as the ``quarter resolution'' versions of $\ell$, $\alpha$, and $\beta$. Then by extension of \eqref{eq:reverse-rgb} we have the relation
\begin{align}\label{eq:quarter-rgb}
w_{\scriptscriptstyle LL}^x(\vec{n}) \approx 
\left[ \begin{array}{ccc} 1 & 2 & 1\vspace{0.1cm}\\ 1 & 0 & -1\vspace{0.1cm}\\ 1 & -2 & 1 \end{array} \right]
\left[ \begin{array}{c} w_{\scriptscriptstyle LL}^y(\vec{n}) \vspace{0.1cm}\\ v_s^{y}(\vec{n})\vspace{0.1cm}\\ w_{\scriptscriptstyle HH}^y(\vec{n}) \end{array} \right]  
\end{align} 
(replace $v_s^y$ by $v_s^y/2ka$ if lossy). That is, we approximately recover a quarter resolution color image $w_{\scriptscriptstyle LL}^{x}(\vec{n})$ directly from the decorrelated one-level wavelet transform coefficients.

We propose to apply color correction, white balance, and gamma correction to the quarter resolution color image $w_{\scriptscriptstyle LL}^{x}$. See \fref{fig:cfa-pipeline}. These steps match the processing that takes place in the digital camera processing pipeline (as described in \sref{sec:camera-processing}), rendering a quarter resolution version of the final color image $\vec{x}_{gc,wb,cc}$ that the digital camera would yield (i.e. the image that we are after). See \fref{fig:rgb-quarter-rgb}. Hence, we posit that a compression method designed to minimize the distortion of the color corrected, white balanced, gamma corrected low-resolution color image would also maximize the quality of the high-resolution color image subsequently reconstructed from the decompressed CFA sampled image. For instance, \fref{fig:cfa-pipeline} illustrates a subsequent JPEG2000 compression, where we convert the gamma corrected color image to luma/chroma components, perform $N$-level wavelet transform and quantize before encoding the bits. 

Finally, when recovering the quarter resolution color image in \eqref{eq:quarter-rgb}, a few coefficients can take on negative values. Thresholding them to zero would introduce additional distortion, which is unattractive. Instead, we keep the absolute value of $w_{\scriptscriptstyle LL}^{x}$, encoding the sign bits separately. The binary image of sign bits is encoded by the standard encoder, which added about $0.004$ bits per pixel on average in our tests. The advantages to the compression scheme in \fref{fig:cfa-pipeline} is confirmed by the experiments in \sref{sec:pipeline-results}.

\begin{figure*}[t]
\centering
\includegraphics[scale=0.5]{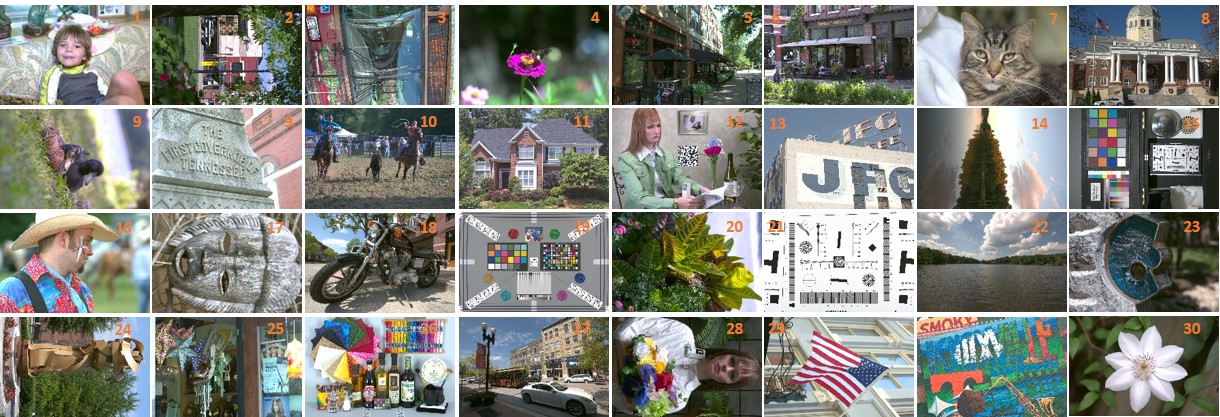}\\
\footnotesize{(a) Test images of Nikon D810} \vspace{0.2cm}\\
\includegraphics[scale=0.5]{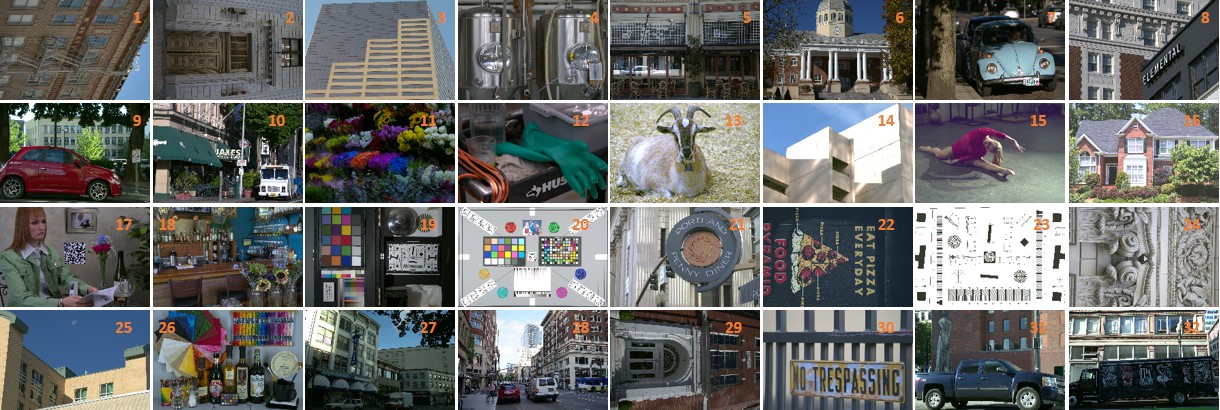} \\
\footnotesize{(b) Test images of Sony Alpha a7R II}  \vspace{0.2cm}
\caption{Raw sensor images used in our experiments. Source is \url{www.imaging-resource.com}, reproduced with a written permission.}
\label{fig:testImages}
\end{figure*}

%\begin{figure*}[t]
%\begin{minipage}{0.49\linewidth}
%  \centering
%  \centerline{\includegraphics[scale=1.0]{nikon_test_data_rs}}
%  \centerline{\footnotesize{(a)} Test images of Nikon D810}\medskip
%\end{minipage}
%\begin{minipage}{0.49\linewidth}
%  \centering
%  \centerline{\includegraphics[scale=1.0]{sony_test_data_rs}}
%  \centerline{\footnotesize{(b)} Test images of Sony Alpha a7R II}\medskip
%\end{minipage}
%\caption{Raw sensor images used in our experiments. Source is \url{www.imaging-resource.com}, reproduced with a written permission.}
%\label{fig:testImages}
%\end{figure*}
\begin{figure*}[t]
\begin{minipage}{0.16\linewidth}
  \centering
  \centerline{\includegraphics[scale=1.0]{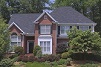}}
  \centerline{\footnotesize{(a)}}\medskip
\end{minipage}
\begin{minipage}{0.16\linewidth}
  \centering
  \centerline{\includegraphics[scale=1.0]{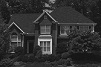}}
  \centerline{\footnotesize{(b)}}\medskip
\end{minipage}
\begin{minipage}{0.16\linewidth}
  \centering
  \centerline{\includegraphics[scale=1.0]{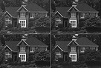}}
  \centerline{\footnotesize{(c)}}\medskip
\end{minipage}
\begin{minipage}{0.16\linewidth}
  \centering
  \centerline{\includegraphics[scale=1.0]{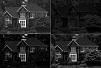}}
  \centerline{\footnotesize{(d)}}\medskip
\end{minipage}
\begin{minipage}{0.16\linewidth}
  \centering
  \centerline{\includegraphics[scale=1.0]{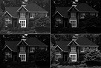}}
  \centerline{\footnotesize{(e)}}\medskip
\end{minipage}
\begin{minipage}{0.16\linewidth}
  \centering
  \centerline{\includegraphics[scale=1.0]{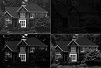}}
  \centerline{\footnotesize{(f)}}\medskip
\end{minipage}
\caption{Comparison of one-level decomposition. (a) RGB. (b) CFA. (c) Demux. (d) MSST. (e) Mallat. (f) Proposed. }
\label{fig:comparisons}
\end{figure*} 

\section{Experimental Results}
\label{sec:results}

\subsection{Setup}
We verified our proposed method using $64$ raw sensor images of Nikon D810 and Sony Alpha a7R II as shown in Fig.~\ref{fig:testImages}. These cameras are especially challenging to work with because they lack the optical lowpass filters.~(We did not consider the low-resolution Kodak and IMAX images typically used in demosaicking and compression studies, as the resolution of typical modern digital cameras far exceeds them. They also poorly approximate the sensor data, prior to the camera pipeline.) In the proposed method, the transformed images were coded using the standard JPEG2000 codec, {\it openJPEG}. The level of wavelet transformation was set to $5$, which is the default setting of {\it openJPEG}. The size of a code block was set to $64 \times 64$ as a default value. The raw sensor data was cropped to the size $4436 \times 6642$ (Nikon) and $4788 \times 7200$ (Sony) because of the size limitation of {\it openJPEG} engine. 

For comparison, we decomposed the color images and the raw sensor images using LeGall $5/3$ wavelet transform for lossless coding and Daubechies $9/7$ wavelet transform for lossy coding of JPEG2000 in different configurations. They are: the wavelet transform on the color images reconstructed in a typical camera processing pipeline~(denoted as ``RGB'' in Table~\ref{tab:bitrates}), the wavelet transform on the CFA sampled images treated as gray images~(``CFA''), the wavelet transform on the demultiplexed color images at low resolution~(``Demux''), the macropixel spectral-spatial transformation~(``MSST'') on the CFA sampled images~\cite{Malvar12}, and the Mallat wavelet packet transformation on the CFA sampled images~(``Mallat'')~\cite{Zhang06}. see Fig.~\ref{fig:comparisons}.

\subsection{Lossless Compression of CFA sampled images}
The compression results are tabulated in Table~\ref{tab:bitrates} in terms of bits per pixel~(bpp). Our proposed method improves coding efficiency by about $7\%$~(Nikon) and $15\%$~(Sony) relative to the standard JPEG2000 coding. Since the total number of pixel components being coded is three times than that of a CFA sampled image in the standard JPEG2000 coding, the RGB encoding is obviously the least efficient. The large coding gain of Sony camera stems from the fact that the dynamic range of raw data was close to $2^{13}$ while the processed color image exceeded $2^{13}$ requiring an extra bit of representation. Typically, the dynamic range of the color images are increased through the camera processing pipeline if raw data are dark, and so this is not entirely unexpected. Relative to directly applying JPEG2000 lossless compression to CFA sampled images, we gained $9.08\%$~(Nikon) and $5.86\%$~(Sony). This can be explained by Fig.~\ref{fig:DWTofYandCFA}(b), which shows that the {\it HL}, {\it LH}, and {\it HH} subbands are dominated by non-sparse chrominance components. Relative to the demultiplexing method, the MSST and the Mallat wavelet transform compression schemes improved by about $5\%$~(Nikon) and $0.7\%$~(Sony). The proposed scheme improved further by $0.71\%$~(Nikon) and $0.40\%$~(Sony) relative to the MSST; and $0.36\%$~(Nikon) and $0.41\%$~(Sony) relative to the Mallat wavelet scheme. The bpps in Table~\ref{tab:bitrates} validate our theoretical analysis in Section~\ref{sec:method} that the decorrelated Mallat wavelet packet transform on CFA sampled images reduces bit requirement.

\begin{table*}
\caption{The comparison of bits per pixel~($bpp$) for different decomposition methods. The bit depth of the raw sensor readout for both cameras is $14$ bits. The coding gain of the proposed method with respect to the other methods is computed as $(bpp\:of\:other\:methods - bpp\:of\:our\:method)/bpp\:of\:other\:methods$ in percentage. The total number of pixel components being coded is three times than that of a CFA sampled image in the standard JPEG2000 coding~({\it RGB}).} \label{tab:bitrates}
\begin{center}
{\renewcommand{\arraystretch}{1.2}
\begin{tabular}{@{}|@{}c@{}| c|c|c|c|c|c| c|c|c|c|c|c|@{}}
\hline
\multirow{2}{*}{Images} & \multicolumn{6}{c|}{Nikon D810} &  \multicolumn{6}{c|}{Sony Alpha a7R II} \\ \cline{2-13}
& {\it RGB} & {\it CFA} & {\it Demux} & {\it MSST} & {\it Mallat} & {\it Ours} 
& {\it RGB} & {\it CFA} & {\it Demux} & {\it MSST} & {\it Mallat} & {\it Ours} \\ \hline \hline
1  & 8.362  & 8.814  & 8.605    & \textbf{8.198}    & 8.206 & 8.204 
   & 7.067  & 6.923  & 6.496    & 6.351   & 6.336	& \textbf{6.296} \\ \hline
2  & 7.186  & 6.814  & 6.564    & 6.187   & 6.167	& \textbf{6.146} 
   & 7.461  & 7.010	 & 6.666	& 6.591   & 6.599	& \textbf{6.579} \\ \hline
3  & 7.688  & 7.894	 & 7.689	& 7.287	  & 7.264	& \textbf{7.224}
   & 7.501  & 6.816	 & 6.348	& 6.366	  & 6.359	& \textbf{6.343} \\ \hline
4  & 7.925  & 7.334	 & 7.142	& \textbf{6.816}	& 6.843 & 6.842 
   & 7.131  & 6.512	 & 6.127	& 6.055	  & 6.053	& \textbf{6.035} \\ \hline
5  & 7.714  & 7.646	 & 7.547	& 7.132	  & 7.113	& \textbf{7.048} 
   & 7.144  & 6.565	 & 6.242	& 6.157	  & 6.131	& \textbf{6.105} \\ \hline
6  & 7.658  & 8.037	 & 7.821	& 7.411	  & 7.384	& \textbf{7.335}
   & 7.696  & 7.177	 & 6.958	& 6.913	  & 6.882	& \textbf{6.827} \\ \hline
7  & 8.221  & 8.840	 & 8.476	& 8.064	  & 8.064	& \textbf{8.048} 
   & 6.827  & 6.101	 & 5.755	& 5.686	  & 5.698	& \textbf{5.668} \\ \hline
8  & 7.949  & 8.591	 & 8.435	& 7.970	  & 7.940	& \textbf{7.912}
   & 7.098  & 6.667	 & 6.329	& 6.242	  & 6.236	& \textbf{6.212} \\ \hline
9  & 7.481  & 7.440	 & 6.989	& 6.654	  & 6.662	& \textbf{6.652}
   & 7.174  & 6.150	 & 5.738	& \textbf{5.674}	& 5.718	& 5.680 \\ \hline
10 & 8.865  & 8.155	 & 8.050	& 7.582	  & 7.528	& \textbf{7.499}
   & 7.395  & 6.817	 & 6.563	& 6.464	  & 6.471	& \textbf{6.429} \\ \hline
11 & 8.278	& 8.088	 & 7.933	& 7.530	  & 7.495	& \textbf{7.476}
   & 8.937  & 7.979	 & \textbf{7.790}     & 7.802   & 7.826 & 7.811 \\ \hline
12 & 9.544  & 8.841	 & 9.003	& 8.426	  & 8.329   & \textbf{8.263}
   & 6.575  & 6.222	 & 5.609	& \textbf{5.601}    & 5.618 & 5.603 \\ \hline
13 & 6.579  & 8.579	 & 7.735	& 7.426	  & 7.405	& \textbf{7.390} 
   & 11.139 & 8.791	 & \textbf{8.598}  & 8.640 & 8.681 & 8.690 \\ \hline
14 & 7.594	& 8.970	 & 8.567	& 8.196	  & 8.156	& \textbf{8.122}
   & 8.006  & 6.440	 & 6.048	& 5.995	  & 6.005	& \textbf{5.989} \\ \hline
15 & 7.806	& 8.106	 & 7.719	& 7.362	  & 7.349	& \textbf{7.335}
   & 7.450  & 8.189	 & 7.861	& 7.840	  & 7.824	& \textbf{7.817} \\ \hline
16 & 7.198	& 7.027	 & 6.847	& 6.505	  & 6.502	& \textbf{6.474}
   & 9.412  & 7.480	 & 7.482	& 7.347	  & 7.268	& \textbf{7.205} \\ \hline
17 & 8.583	& 8.512	 & 8.252	& 7.872	  & 7.860	& \textbf{7.843}
   & 7.241  & 7.946	 & 7.419	& 7.387	  & 7.341	& \textbf{7.319} \\ \hline
18 & 10.178	& 9.180	 & 9.128	& 8.670	  & 8.650	& \textbf{8.634}
   & 8.513  & 7.344	 & 7.095	& \textbf{7.083}    & 7.103 & 7.090 \\ \hline
19 & 7.198  & 7.375	 & 7.245	& 6.783	  & 6.739	& \textbf{6.699}
   & 7.301  & 6.207	 & 5.987	& 5.966	  & 5.966	& \textbf{5.944} \\ \hline
20 & 8.000	& 8.826	 & 8.334	& 7.937	  & 7.939	& \textbf{7.907} 
   & 8.272  & 7.843	 & 7.403	& 7.434	  & 7.432	& \textbf{7.393} \\ \hline
21 & 6.962	& 7.676	 & 7.029	& 6.698	  & 6.709	& \textbf{6.688}
   & 7.535  & 6.820	 & 6.375	& 6.319	  & 6.333	& \textbf{6.315} \\ \hline
22 & 7.949	&8.589	 & 7.638	& 7.270	  & \textbf{7.264} &7.277
   & 7.773  & 6.161	 & 6.087	& 5.947	  & 5.973	& \textbf{5.931} \\ \hline
23 & 6.733	& 7.467	 & 7.104	& 6.683	  & 6.651	& \textbf{6.628}
   & 8.183  & 7.343	 & 6.851	& 6.850	  & \textbf{6.845} & 6.845 \\ \hline
24 & 7.878	& 7.721	 & 7.784	& 7.321	  & 7.229	& \textbf{7.184}
   & 8.537  & 6.937	 & 6.670	& 6.540	  & 6.537	& \textbf{6.516} \\ \hline
25 & 7.601	& 7.548	 & 7.485	& 7.013	  & 6.908	& \textbf{6.871} 
   & 7.685  & 6.661	 & 6.181	& 6.143	  & 6.109	& \textbf{6.093} \\ \hline
26 & 7.327	& 7.524	 & 7.076	& 6.752	  & 6.764	& \textbf{6.740} 
   & 8.024  & 7.339	 & 7.151	& 7.073	  & 7.0268  & \textbf{6.988} \\ \hline
27 & 8.203	& 9.144	 & 8.979	& 8.461	  & 8.414	& \textbf{8.373}
   & 7.241  & 6.659	 & \textbf{6.271}	  & 6.296	& 6.337	& 6.290 \\ \hline
28 & 7.637	& 8.119	 & 7.955	& 7.516	  & 7.471	& \textbf{7.428}
   & 7.122  & 6.762	 & 6.327	& \textbf{6.302}	& 6.376	& 6.334 \\ \hline
29 & 7.339	& 7.460	 & 7.218	& 6.793	  & 6.758	& \textbf{6.734} 
   & 7.738  & 6.849	 & 6.533	& 6.492	  & 6.507	& \textbf{6.470} \\ \hline
30 & 7.422	& 7.500	 & 7.026	& 6.658	  & 6.656	& \textbf{6.641}
   & 8.274  & 6.422	 & 6.085	& \textbf{6.073}	& 6.087	& 6.077 \\ \hline
31 & 10.273	& 9.403	 & 9.253	& 8.875	  & 8.809	& \textbf{8.767}
   &  7.337 & 6.829	 & 6.489	& 6.441	  & 6.403	& \textbf{6.363} \\ \hline
32 & 8.647	& 7.270	 & 6.951	& 6.631	  & 6.636	& \textbf{6.624} 
   & 7.465  & 6.746	 & 6.430	& 6.406	  & 6.413	& \textbf{6.380} \\ \hline
\hline
Average  & 7.937 & 8.078 & 7.800 & 7.396 & 7.371 & \textbf{7.344} 
         & 7.758 & 6.960 & 6.624 & 6.577 & 6.578 & \textbf{6.551} \\ \hline
Our      & \multirow{2}{*}{7.47\%} & \multirow{2}{*}{9.08\%} & \multirow{2}{*}{5.83\%} & \multirow{2}{*}{0.71\%} 
    & \multirow{2}{*}{0.36\%} & \multirow{2}{*}{$-$} 
         & \multirow{2}{*}{15.6\%} & \multirow{2}{*}{5.86\%} &\multirow{2}{*}{1.10\%} & \multirow{2}{*}{0.40\%}
    & \multirow{2}{*}{0.41\%} & \multirow{2}{*}{$-$} \\ 
Gain     & & & & & & & & & & & & \\ \hline
\end{tabular} }
\end{center}
\end{table*}

\begin{figure*}[t]
\begin{minipage}{0.45\linewidth}
\centering
\includegraphics[scale=0.35]{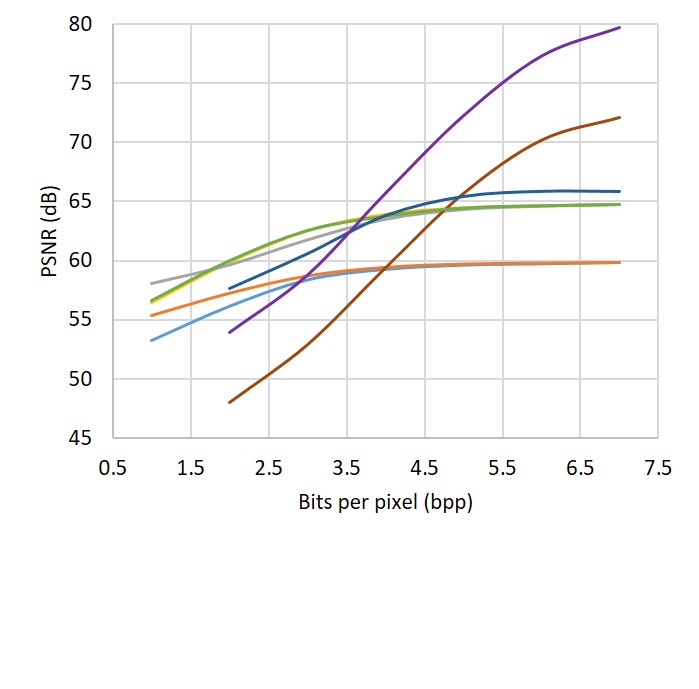}\\
{\footnotesize (a) Nikon D810}
\vspace{0.2cm}
\end{minipage}
\begin{minipage}{0.45\linewidth}
\centering
\includegraphics[scale=0.35]{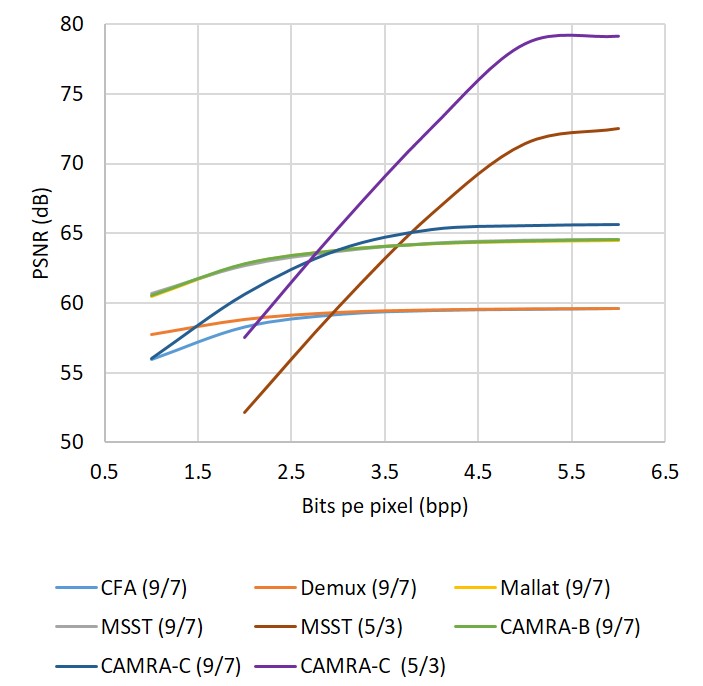}\\
{\footnotesize (b) Sony Alpha a7R II}
\vspace{0.2cm}
\end{minipage}
\vspace{0.2cm}
\caption{The comparison of lossy coding schemes. The decorrelated Mallat packet wavelet structure in \sref{sec:lossy} and the further lossy docing schme in \sref{sec:alternative} are denoted as {\it Ours} and {\it Ours-A}, respectively. The Daubechies $9/7$ wavelet transform and the LeGall $5/3$  were used for the first level in {\it Ours-A (9/7)} and in {\it Ours-A (5/3)}. The standard coding scheme with MSST decomposition and the lossy coding scheme in~\cite{Malvar12} are denoted as {\it MSST} and {\it MSST-A}. \label{fig:lossy-perf-psnr}}
\end{figure*}
\begin{figure*}
\begin{minipage}{0.5\linewidth}
\end{minipage}
\begin{minipage}{0.5\linewidth}
\centering
\includegraphics[scale=0.3]{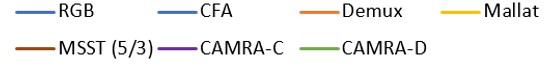}
\end{minipage}
\\
\begin{minipage}{0.5\linewidth}
\centering
\includegraphics[scale=0.4]{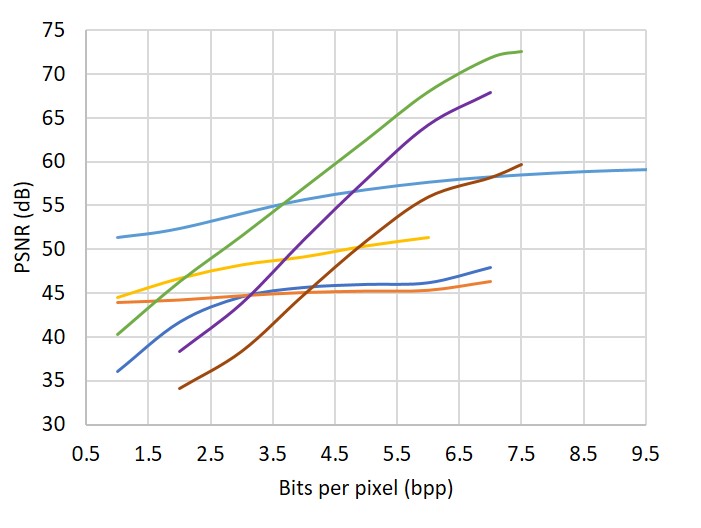} \\
\footnotesize{(a) Nikon, LSLCD}
\end{minipage}
\begin{minipage}{0.5\linewidth}
\centering
\includegraphics[scale=0.4]{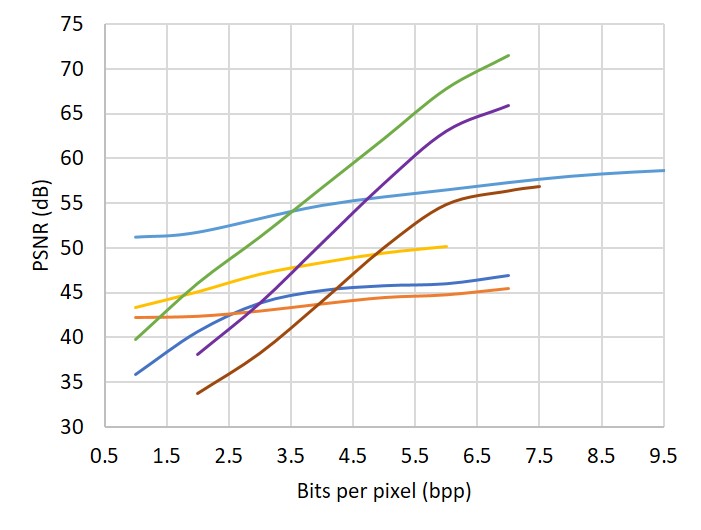}\\
\footnotesize{(b) Nikon, PSDD}
\end{minipage}
\\
\begin{minipage}{0.5\linewidth}
\centering
\includegraphics[scale=0.4]{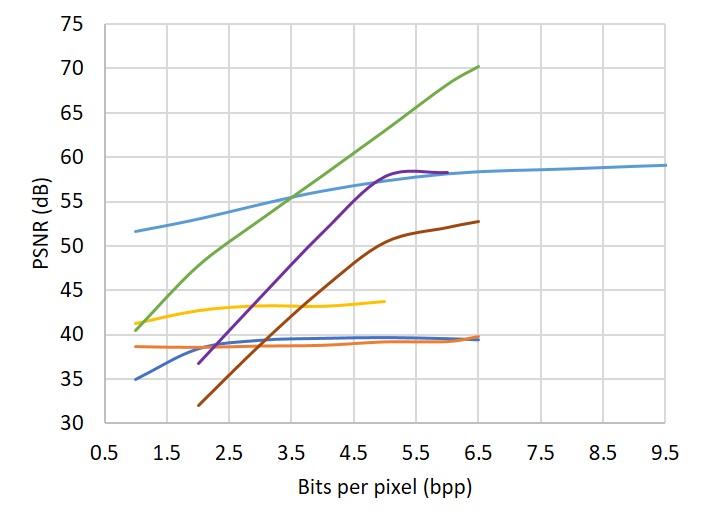} \\
\footnotesize{(c) Sony, LSLCD}
\end{minipage}
\begin{minipage}{0.5\linewidth}
\centering
\includegraphics[scale=0.4]{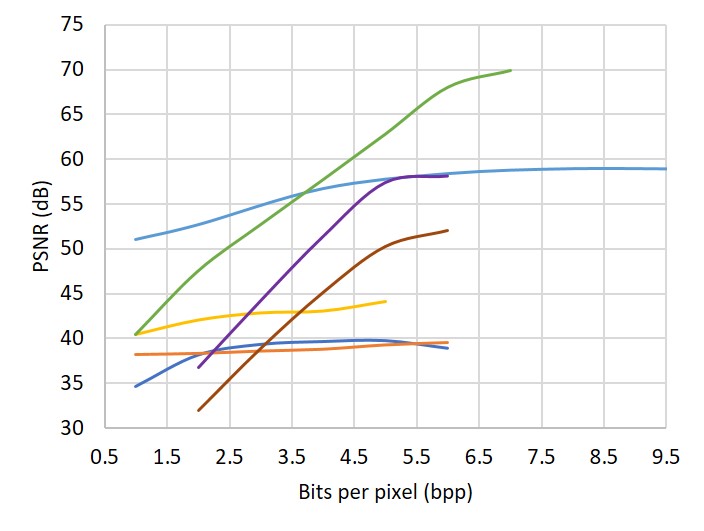}\\
\footnotesize{(d) Sony, PSDD}
\end{minipage}

\caption{Performance comparisons of the color images. The pipeline optimization scheme in \sref{sec:pipeline} denotes as {\it Ours-A-P}.}
\label{fig:pipeline-perf}
\end{figure*}

\subsection{Lossy Compression of CFA sampled images}
\label{sec:lossy-results}
The performance of the proposed decorrelated Mallat packet wavelet scheme in \sref{sec:lossy} and the further optimization scheme in \sref{sec:alternative} were evaluated on the rate-distortion curves in \fref{fig:lossy-perf-psnr}. In the curves, we measured the average peak signal-to-noise ratio~(PSNR) of the reconstructed test CFA sampled images, relative to the original CFA sampled images. We applied the standard JPEG2000 lossy coding scheme to the comparing methods~(including the method in \cite{Zhang06}, which was proposed for lossless coding only). In this experiment, bitrates were controlled by quantization step. We disabled the rate-controller in openJPEG because it confounds the distortion analysis. 

The alternative lossy coding scheme in \sref{sec:alternative} outperformed at higher bitrates as predicted in the analysis of  \sref{sec:method}. At lower bitrates, the decorrelated Mallat packet wavelet scheme slightly improved the coding efficiency. This demonstrates that round-off error is predominated at higher bitrates, and quantization error increases as bitrates decrease.

\subsection{Lossy Compression with Camera Processing Pipeline}
\label{sec:pipeline-results}
Next, we investigate the effects of distortions introduced by the lossy compression on the camera processing pipeline. For this study, we considered two versions of the camera processing pipeline---one using least square luma-chroma demultiplexing~(LSLCD) demosaicking~\cite{Jeon11} and the other using posterior sparsity-directed demosaicking~(PSDD)~\cite{Korneliussen14}. 

In \fref{fig:pipeline-perf}, the quality of the color images reconstructed from the decompressed CFA images using digital camera processing pipeline are evaluated in terms of PSNR. For the reference images for PSNR, we applied the same camera processing pipeline to the original CFA image, since it is the best possible image we can obtain without compression.

We report the rate-distortion curves in \fref{fig:pipeline-perf}. The performance of compressing the full RGB image rendered by the camera processing pipeline was better than the existing CFA image compression technique. It suggests that the distortions introduced by lossy compression scheme indeed do propagate through the camera processing pipeline. By contrast, at modest to high bitrate, the proposed camera processing pipeline-aware compression algorithm clearly outperforms the full RGB and the existing CFA image compression schemes.

\subsection{Computational Complexity}
The total processing time is dominated by the wavelet transformation and the variable length coding. We only require wavelet transform of a CFA sampled image at the first level, which is one-third the complexity of the JPEG2000 for color images. The decorrelation requires only an addition and a subtraction, which is a negligible overhead. The subsequent $N$-dimensional wavelet transform is applied three times, comparable to JPEG2000 for color images (after the first level wavelet decomposition); and requires $3/4$ complexity relative to other decomposition schemes in \fref{fig:comparisons}. The number of wavelet coefficients that are encoded by the variable length encoder is one-third of the JPEG2000. Hence, we conclude that the proposed method is substantially less complex than the conventional JPEG2000 and slightly less than other compression schemes.  

\section{Conclusions}
\label{sec:conclusion}
We proposed a novel lossless and lossy compression schemes for CFA sampled images based on the wavelet analysis of CFA sampling. The analysis proved both theoretically and experimentally that the wavelet coefficients of CFA sampled images are highly correlated. Consequently, we developed the lossless and lossy compression schemes to further sparsify the correlated coefficients. The experimental results verified that the proposed scheme improve coding efficiency relative to the compression of color images. Furthermore, we designed the camera processing pipeline~(CAMRA) that minimizes the error of the color images reconstructed from the uncompressed CFA sampled data by the subsequent camera processing pipeline. The experimental results using actual sensor data verified that the proposed decomposition schemes improves coding efficiency relative to the standard color image compression schemes as well as the state-of-the-art compression schemes for CFA sampled images.

%\appendices
%\section{Proof of the First Zonklar Equation}
%Appendix one text goes here.
\bibliographystyle{IEEEtran}
\bibliography{IEEEabrv,refs}

% use section* for acknowledgment
\section*{Acknowledgment}
We thank the staff at Image Resource for giving us the permission to reproduce the raw sensor images used in this study. We also thank openJPEG for the encoding scheme we used to produce results.

% that's all folks
\end{document}